\documentclass[runningheads]{llncs}
\usepackage[T1]{fontenc}
\usepackage{graphicx}
\usepackage{hyperref}
\usepackage{color}

\usepackage{amsmath}
\usepackage{tikz}
\usetikzlibrary{matrix}
\usepackage[linesnumbered,ruled,vlined]{algorithm2e}
\SetKwInput{KwInput}{Input}
\SetKwInput{KwOutput}{Output}
\usepackage{multirow}

\begin{document}
\title{Integrating Inverse and Forward Modeling \\  for Sparse Temporal Data from Sensor Networks}
\titlerunning{Inverse and Forward Modeling for Sparse Temporal Data}
%
\author{Julian Vexler\inst{1}\orcidID{0000-0002-8336-1447} \and
Björn Vieten\inst{2}\and
Martin Nelke\inst{3}\and \\
Stefan Kramer\inst{1}\orcidID{0000-0003-0136-2540}}
\authorrunning{J. Vexler et al.}
%
\institute{Johannes Gutenberg University Mainz, Staudingerweg 9, 55218 Mainz, Germany
\email{j.vexler@gmx.net,kramer@informatik.uni-mainz.de}\and
Fraport AG, 60547 Frankfurt am Main, Germany
\email{B.Vieten@Fraport.de}\and
MIT GmbH, Pascalstraße 69, 52076 Aachen, Germany
\email{Martin.Nelke@mitgmbh.de}}
\maketitle              
\begin{abstract}
We present CavePerception, a framework for the analysis of sparse data from sensor networks that incorporates elements of inverse modeling and forward modeling. By integrating machine learning with physical modeling in a hypotheses space, we aim to improve the interpretability of sparse, noisy, and potentially incomplete sensor data. The framework assumes data from a two-dimensional sensor network laid out in a graph structure that detects certain objects, with certain motion patterns. Examples of such sensors are magnetometers. Given knowledge about the objects and the way they act on the sensors, one can develop a data generator that produces data from simulated motions of the objects across the sensor field. The framework uses the simulated data to infer object behaviors across the sensor network. The approach is experimentally tested on real-world data, where magnetometers are used on an airport to detect and identify aircraft motions. Experiments demonstrate the value of integrating inverse and forward modeling, enabling intelligent systems to better understand and predict complex, sensor-driven events.

\keywords{Inverse modeling  \and Forward modeling \and Sensor networks \and Magnetometers \and Aviation.}
\end{abstract}
\section{Introduction}
\begin{figure}[t]
    \centering
    \includegraphics[width=0.45\textwidth]{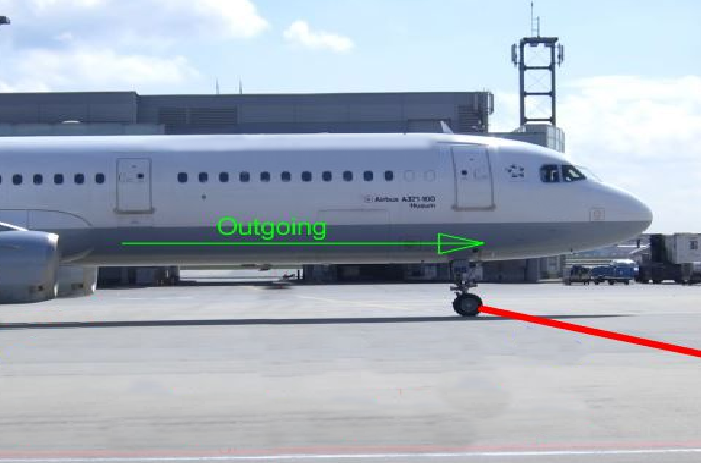}
    \includegraphics[width=0.45\textwidth]{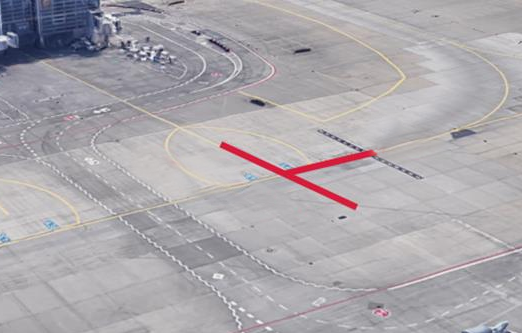}
    \caption{Left: An aircraft leaving a parking area, while trespassing the perpendicular bus line. Right: Sensor field set-up with one bus line along the yellow center guiding line (68 sensors) and one bus line perpendicular to it (53 sensors).
    }
    \label{fig:test_field}
\end{figure}
Sensor networks can collect an almost arbitrary amount of multivariate data in a fraction of a second. However, sensors only measure the effect or outcome of physical phenomena. The true parameters that cause these outcomes are usually unknown.   
\emph{Inverse modeling} is the approach to estimate the unknown causal parameters from the available observations. In other words, models are trained with the obtained data, in order to estimate the unknown parameters. Machine Learning (ML) approaches are the state of the art for making inferences from data. However, they frequently achieve their peak performance when the true labels are known and they can be trained in a supervised fashion. 
In contrast to inverse modeling, \emph{forward modeling} uses well-defined parameters to generate synthetic data from simulations in, usually, controlled environments. 
In this work, we present CavePerception, a framework for the analysis of sparse data from sensor networks that incorporates both inverse modeling and forward modeling. We assume a two-dimensional sensor network, laid out in a graph structure, where the sensors deliver multi-variate data about the surrounding area. The data is influenced by objects of certain categories, which are able to move in different directions at certain speeds and angles. The framework is able to handle any stationary-installed sensors, as long as they deliver time-series data. 
In this work, we consider \emph{magnetometers}, which can be used to measure the distortion of the Earth's magnetic field by metallic objects, components, and external magnetic fields. While the measurements of the sensors themselves are not giving information about the category of the object or its motion vector, knowledge about the overall setting can be used to derive valid options for object categories and motion vector assignments. This knowledge may be used to develop a synthetic data generator, which mimics the environment and sensor set-up, to simulate a motion for each of the valid options. 
In CavePerception, we integrate ML-based inverse modeling and forward modeling to develop multiple hypotheses about the events detected by the sensors. The inverse modeling part predicts a category and a motion vector for the observation, while the forward modeling part aims at improving the predictions and filling in missing forecasts, in case the inverse modeling was not able to predict a parameter. 
Our framework is tested on real-world data, where 121 magnetometers are used at Frankfurt airport in Germany, to detect and identify aircraft motions. The sensor field is shown in Figure \ref{fig:test_field}. 
The experiments demonstrate the value of integrating inverse and forward modeling, in comparison to conventional ML approaches. 
In future work, CavePerception shall be integrated into novel magnetometer-based traffic surveillance systems on airport aprons, which aim to replace outdated systems, as these often fail to prevent hazardous situations\footnote{\href{https://www.aerotime.aero/articles/air-europa-b737-collides-with-condor-b757-at-palma-de-mallorca-airport}{https://www.aerotime.aero/articles/air-europa-b737-collides-with-condor-b757-at-palma-de-mallorca-airport}}. 

In summary, the main contributions of this paper are: (i) CavePerception, a framework for object classification and motion estimation from sensory networks, (ii) a combination of inverse and forward modeling, for an improved recognition of events, (iii) a solution to handle sparse temporal sensor data, and (iv) the validation of the framework in a real-world scenario.

\section{Related Work}
Inverse modeling is currently the predominant paradigm in ML and Data Mining (DM). 
Work on the integration of forward and inverse modeling, however, 
is currently found mainly outside ML and DM. A paper from applied physics  \cite{RobertsHedayati2021}  presents a deep learning approach to the forward and inverse designs of plasmonic metasurface structural color. Another applied physics paper  describes a deep learning framework for forward and inverse problems in time-domain electromagnetics \cite{Huetal2021}. 
Closer in spirit are robotics approaches for tracking and prediction based on deep forward and inverse perceptual models \cite{Lambertetal2018}. Deriving data from simulations of theoretical models and then training classification, regression, and ranking models is quite common in ML applied to particle physics \cite{Koeppeletal2021}. One has to note that the integration of forward and inverse modeling is similar to abduction \cite{Kakas2017}, a currently underrepresented topic in DM \cite{Wickeretal2015}.

In case of traffic surveillance on airport grounds, there is no actual standard for sensor systems for the tasks of object detection, speed estimation, tracking, and classification. Consequently, different technological systems are explored in the literature, where most of them focus on trajectory optimization during flight \cite{ref:tian2020}, mid-air classification of aircraft using radar \cite{ref:xia2022} or aircraft classification based on top-view images \cite{ref:gao2022,ref:azam2021}. 
However, in case of surface traffic surveillance on airports, aircraft need to be classified on the ground. Hence, visual-based technologies may struggle with obstacles and visually bad weather conditions like heavy rain, snow or fog \cite{ref:zhang2023}. On the other hand, magnetometers are not affected as they measure the Earth's magnetic field. They may also be installed directly below the surface of the holding point positions that need to be monitored, functioning as on-site sensors. Consequently, magnetometers have a high potential to be a part of traffic surveillance systems on airport grounds.   

\section{Simulation-Assisted Forward Modeling for Sparse Temporal Sensor Data}
Our solution to the challenge of object detection, motion estimation, and classification of a moving object based on multivariate time-series data, is mirrored in our framework, named CavePerception. Inspired by Plato's allegory of the cave, where humans only see shadows of objects, which are inaccurate representations of ideal objects, we interpret our problem as follows: The sensors measurements are reflections of object motions, yet, they are not able to accurately represent the objects themselves. Consequently, the goal of CavePerception is to perceive real-world observations as imperfect representations of forwardly modeled synthetic patterns that are generated from known objects. Despite the imprecision in the data, the matching between observed and well-defined patterns shall reveal the true object.

A top-level view of the processing pipeline of CavePerception is shown in Algorithm \ref{alg:caveperc} and explained in detail below. The inverse modeling aims to classify an object and estimate its motion vector based on the information retrieved from the sensory data and geometric domain knowledge. This part of the framework is based on previous work \cite{ref:vexler2023motion,ref:vexler2023classifying}, briefly summarized below. The forward modeling and hypotheses matching parts are the novel aspects of CavePerception. 
Overall, the aim is to improve the results from inverse modeling, by assisting the data analysis with well-defined, knowledge-infused simulations.

\begin{algorithm}[t]
\SetAlgoLined
\caption{CavePerception}
\label{alg:caveperc}
\KwInput{Initial hypotheses set $\mathcal{H}_0$}
\KwOutput{Predicted hypotheses $\Tilde{\mathcal{H}}$}
\KwData{Detected real-world event $E$}
// Inverse Modeling \\
\Indp $\overrightarrow{v} \longleftarrow$ estimate motion vector \\
$\phi(E) \longleftarrow$ classify object category \\
$\mathcal{H}_{real} \subseteq \mathcal{H}_0,$ inversely modeled hypotheses\\
\Indm
// Forward Modeling\\
\Indp $\mathcal{H}_{sim} \longleftarrow$ simulation hypotheses generated by a synthetic data generator\\
\Indm
// Inverse and Forward Modeling Matching\\
\Indp $\tilde{\mathcal{H}} \in \mathcal{H}_{real} \cap \mathcal{H}_{sim},$ retrieve the most likely hypotheses (see Algorithm \ref{alg:hypo_match})\\
\Indm
\end{algorithm}

\paragraph{Inverse Modeling} 
In the context of signal processing, it is often required to first employ a signal detection algorithm, before applying subsequent algorithms. 
Let $F=\{s_1,...,s_k\}$ denote a sensor field, where each sensor $s_i \in F$ delivers a multivariate vector reading. In case of 3-axis magnetometers, it is a three-dimensional vector of the Earth's magnetic field at each time point $t$: $s_i^t=\{X_i^t,Y_i^t,Z_i^t\}$.
We are only interested in sensor signals, when an object $\rho$ is present. Hence, an \textit{event} detection algorithm based on a Z-score approach\footnote{\href{https://stackoverflow.com/questions/22583391/peak-signal-detection-in-realtime-timeseries-data/22640362}{https://stackoverflow.com/questions/22583391/peak-signal-detection-in-realtime-timeseries-data/22640362}} is applied, to calculate a moving threshold $\tau(s_i^t)$ for each sensor $s_i \in F$ at time $t$. A sensor $s$ is \textit{activated} if its measurements exceed the threshold: $|s^t| \geq \tau(s^t)$.
Furthermore, we define an \textit{event} $E(t_0,t_k)$ as a sequence of activated sensors over time, from $t_0$ to $t_k$, as:
\begin{equation}
    E(t_0,t_k) := \{s^{t_m}|\forall s \in F, m \in [0,k]: |s^{t_m}| \geq \tau(s^{t_m})\},
\end{equation}
where at least one sensor is activated at each time point $t_m$.
For simplification, event $E(t_0,t_k)$ is abbreviated as $E$.
In real-world scenarios, an event $E$ captures a temporal frame of a moving object $\rho$, where the object activates one or multiple sensors at each time point in $E$. Ideally, the complete object motion is observed in one event $E$. However, depending on the sensor field set-up, the sensor type, and the object itself, a crossing may be not continuously observable by the sensors, leading to a separation into multiple coherent events, e.g., $E_1, E_2,$ and $E_3$.

The inverse modeling approach of CavePerception processes the sensors measurements by applying an object detection algorithm. As soon as an object is detected within an event $E$,
CavePerception aims to reduce an initially user-defined set of hypotheses $\mathcal{H}_0 = \{h_1,h_2,...,h_n\}$. $\mathcal{H}_0$ contains all hypotheses $h_i = (\rho_i,\overrightarrow{v}_i)$ with object $\rho_i$ and motion vector assignments $\overrightarrow{v}_i$ that may be observed in the set up environment, i.e. $\mathcal{H}_0$ is defined based on background knowledge. 
The activated sensors within $E$ are clustered using ST-DBScan \cite{ref:birant2007} to retrieve a first geometrical pattern of the object.
A cluster fusion approach \cite{ref:vexler2023motion} combines clusters to obtain more accurate object representations. Furthermore, a data-driven event fusion algorithm \cite{ref:vexler2023motion} is incorporated, to fuse consecutive events $E_i$ if they were triggered by the same object. Based on the clustering results, the motion vector $\overrightarrow{v}$ of the object $\rho$ is estimated. Finally, a Transformer-based \cite{ref:vaswani2017} classification of the object is performed to categorize the object \cite{ref:vexler2023classifying}. 
At the end of the inverse modeling procedure, CavePerception returns a set of hypotheses 
$\mathcal{H}_{real} \subseteq \mathcal{H}_0$, where $h_i = (\phi(E),\overrightarrow{v}_i) \in \mathcal{H}_{real}$ contains the Transformer's classification $\phi(E)$ and the estimated motion vectors $\overrightarrow{v}_i$. 

\paragraph{Forward Modeling} The idea behind forward modeling is to have a synthetic data generator, which generates data similar to the real-world data. 
Therefore, we make use of geometrical knowledge about the sensor field installation and the geometric pattern of polygonal-shaped objects. The user can then define which objects shall be simulated, how the sensors are positioned in a two-dimensional grid, and specify the motion behaviour of the objects to be simulated. 
Additional parameters like the starting position, the sensor detection range, and the data noise level may be defined. 
Based on the application field, background knowledge is used to perform arbitrarily many simulations of varying complexities to retrieve all simulation hypotheses $\mathcal{H}_{sim}$ that would be expected as valid observations for the considered sensor field. At the same time, a synthetic data set $\mathcal{D}_{S_i}$ is obtained for each simulation $S_i$, or equivalently, each hypothesis $h_i \in \mathcal{H}_{sim}$.


\paragraph{Inverse and Forward Modeling Matching} In real-world domains, a data-driven inverse modeling approach may not always deliver the full set of required or accurate results. 
The idea of the forward modeling approach is to fill the gap of missing information and to improve the derived results by matching realistic simulations with real-world observations. 

After CavePerception's inverse modeling pass, each hypothesis $h \in \mathcal{H}_{real}$ contains two values: $h=(\phi(E),\overrightarrow{v})$, where 
$\overrightarrow{v}=\{d,\nu,\alpha \}$ is a motion vector with motion direction $d$, motion velocity $\nu$, and motion angle $\alpha$. If the inverse pass is able to predict any of these parameters, then all hypotheses $h \in \mathcal{H}_{real}$ have the same value for the respective parameter. 
For example, if $d=right$, then $\forall h=(\phi(E),\overrightarrow{v})\in \mathcal{H}_{real}: \overrightarrow{v}=\{right,\nu,\alpha\}$. 
After the inverse pass, there are four different scenarios: (i) the object was classified and a motion was detected (fully or partially), (ii) the object was classified, but no motion was detected, (iii) a motion was detected, but the object could not be classified, and (iv) there is not enough information to detect a motion or to perform a classification. 

Our forward modeling approach aims to fill the gaps of missing information in $\mathcal{H}_{real}$, including the identification of the specific object type $\rho$, by employing a time-invariant hypotheses matching algorithm within CavePerception. The pseudocode is shown in Algorithm \ref{alg:hypo_match} and explained below.

\paragraph{Hypotheses Matching}
The first step of the algorithm is the matching of the inversely derived hypotheses $\mathcal{H}_{real}$ of event $E$ with the simulated hypotheses $\mathcal{H}_{sim}$. In contrast to $\mathcal{H}_{real}$, the parameters of all hypotheses $h=(\hat{\rho},\overrightarrow{v})\in \mathcal{H}_{sim}$ are known. Therefore, only those simulations are considered where the simulation parameters correspond to the estimated parameters $\phi(E)$ and $\overrightarrow{v}$: 

\begin{equation}
    \mathcal{H}_{red} \longleftarrow \mathcal{H}_{sim} \cap \mathcal{H}_{real}.
\end{equation}

\begin{algorithm}[t]
\SetAlgoLined
\caption{Inverse and Forward Model Matching}
\label{alg:hypo_match}
\KwOutput{Final hypotheses $\tilde{\mathcal{H}}$}
\KwData{Hypotheses derived from real-world measurements $\mathcal{H}_{real}$, Synthetic hypotheses $\mathcal{H}_{sim}$, Real-world observed event $E$}
// Hypotheses matching\\ 
$\mathcal{H}_{red} \longleftarrow \mathcal{H}_{sim} \cap \mathcal{H}_{real}$\\
// Time-invariant hypotheses reduction\\
$B(\mathcal{D}_{red}) \longleftarrow \{B(\mathcal{D}_{S_i})|\forall \mathcal{D}_{S_i} \in \mathcal{D}_{red}\}$, binary activation matrices of matched synthetic datasets\\
$B(E) \longleftarrow$ binary activation matrix of event $E$\\
$Y_{red} \longleftarrow$ initialize empty set for scalar distances\\
\For{each $S_i \in \mathcal{H}_{red}$}{
    $M_{E,S_i} \longleftarrow pdist(B(E),B(\mathcal{D}_{S_i}))$ // see Eq. 5\\ 
    $Y_{red} \longleftarrow Y_{red} \cup \{sdist(M_{E,S_i})\}$   // see Eq. 6 \\
}
$\tilde{\mathcal{H}} \longleftarrow \{S_i \in \mathcal{H}_{red}|sdist(M_{E,S_i}) = \min (Y_{red})\}$ // see Eq. 7 
\end{algorithm}

The hypotheses reduction to $\mathcal{H}_{red}$ results in hypotheses $h_i \in \mathcal{H}_{red}$, where each parameter is known, as the missing information from the inverse modeling in $\mathcal{H}_{real}$ is filled with the available information in $\mathcal{H}_{sim}$. For example, if the category $\phi(E)$ of object $\rho$ is successfully predicted during inverse modeling, then all simulations of object types $\{\rho_1,...,\rho_n\}$ of category $\phi(E)$ are extracted from $\mathcal{H}_{sim}$. This way, not only the categories of the objects are known, but also the specific object types $\hat{\rho}_i$, which were unknown in $\mathcal{H}_{real}$. 
Here, we use the notation $\hat{\rho}_i$ for the predictions of the true type $\rho$ based on the predicted category $\phi(E)$. 
With the hypotheses matching, the set of simulation datasets $\mathcal{D}_S$ is reduced to: 

\begin{equation}
    \mathcal{D}_{red} =\{\mathcal{D}_{S_i} \in \mathcal{D}_S| h_{S_i}=(\hat{\rho},\overrightarrow{v})\in \mathcal{H}_{red}\},
    \label{equ:sim_match}
\end{equation}

where $\mathcal{D}_{red}$ only contains those simulated datasets where the simulations were performed according to the estimated parameter combinations in $\mathcal{H}_{real}$. 

\paragraph{Time-Invariant Hypotheses Reduction}
After the hypotheses matching, it remains to retrieve the most likely hypotheses from $\mathcal{H}_{red}$ to describe the observed event $E$. Therefore, binary activation matrices of the simulated datasets $\mathcal{D}_{red}$ are compared with the binary activation matrix of event $E$. This way, we aim to find an optimal geometry-based match. Hence, each synthetic dataset $\mathcal{D}_{S_i} \in \mathcal{D}_{red}$ is converted from a continuous data representation to a binary activation matrix $B(\mathcal{D}_{S_i})$, where each column represents a sensor and the rows represent the time: 

\begin{equation}
    B_{i,j} = 
    \begin{cases}
        1 &, \text{if } s_j^{t_i} \geq \tau(s_j^{t_i}),\\
        0 &, \text{otherwise}. 
    \end{cases}
\end{equation}

Subsequently, each binary activation matrix $B(\mathcal{D}_{S_i})$ needs to be matched with $B(E)$. The matches are then used to derive dissimilarity scores that describe to which extent a simulation coincides with the observation. The most similar simulation hypotheses are then returned as the most likely explanations for the observed event.  
The matching of the binary activation matrices can be performed with any pair-wise distance measure $pdist$. The matrices themselves, $B(E)$ and $B(\mathcal{D}_{S_i})$, need to have the same number of columns $n_c$ (the number of sensors), but can vary in the number of rows $n_k$ and $n_l$, i.e. $|B(E)|=(n_k,n_c)$ and $|B(\mathcal{D}_{S_i})|=(n_l,n_c)$, as the lengths of different events may vary. 
The pair-wise distance between the two matrices results in a distance matrix: 
\begin{equation}
\begin{aligned}
     M_{E,S_i} & \longleftarrow pdist(B(E),B(\mathcal{D}_{S_i})) \\
            &= (dist(v_i,v_j))_{i,j}, \forall i \in \{1,...,n_k\}, j \in \{1,...,n_l\}, 
\end{aligned}    
\end{equation}
where $v_i$ and $v_j$ are the $i$-th and $j$-th row vectors of matrices $B(E)$ and $B(\mathcal{D}_{S_i})$, respectively.
In this work, we apply the Hamming distance, which counts at how many positions the entries in two vectors differ.
Applied to the sensor activation matrices, it counts how many sensors differ in their activation in event $E$ and simulation $S_i$. 
\begin{figure}[t]
    \centering
    \includegraphics[width=0.45\textwidth]{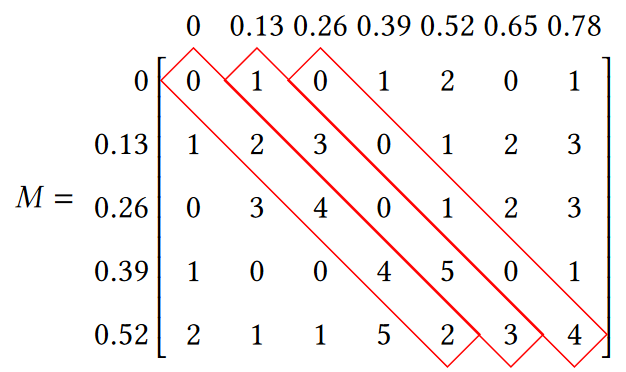}
    \caption{An example of a distance matrix $M$. The rows and columns specify temporal distances to the event start, whereas the entries represent the pairwise distance values. 
    }
    \label{fig:ex_dist}
\end{figure}
Figure \ref{fig:ex_dist} shows an example of such a distance matrix $M$, where the rows and columns represent the temporal distances of both matrices $B(E)$ and $B(\mathcal{D}_{S_i})$.
However, we are interested in a scalar score to describe the dissimilarity between both matrices. Therefore, the diagonals within matrix $M$ are considered (framed in red). Each diagonal is a full-match of observation $E$ within the simulation $S_i$. 
We restrict the diagonals to mirror the full path of the smaller matrix. Hence, each diagonal needs to have $min(n_k,n_l)$ values. If a full match is not required, i.e. both crossings can also match partially, then the length of the diagonals could be set to be less restrictive. We can assure $n_k$ to be smaller than $n_l$ by transposing the resulting matrix, if $n_k > n_l$. 
To obtain a scalar distance $sdist(M_{E,S_i})$, we search for the minimal dissimilarity in the diagonal sums of all diagonals $diag(M) = \{diag_1(M),\allowbreak ...,\allowbreak diag_k(M)\}$:
\begin{equation}\label{equ:sdist}
    sdist(M_{E,S_i}) = min_{i \in \{1,...,|diag(M)|\}} \sum_{v \in diag_i(M)} v.
\end{equation}
Equation \ref{equ:sdist} returns the optimal time-invariant match of event $E$ with simulation $S_i$.
Applying Equation \ref{equ:sdist} to every simulation $S_i \in \mathcal{H}_{red}$, a set of optimal time-invariant matches is obtained and collected in $Y_{red}$.
Finally, Algorithm \ref{alg:hypo_match} concludes the search for the optimal match in the simulations by choosing the simulations with the lowest scalar dissimilarity:
\begin{equation}
    \tilde{\mathcal{H}} \longleftarrow \{S_i \in \mathcal{H}_{red}|sdist(M_{E,S_i}) = \min (Y_{red})\}.
\end{equation}
Notice that synthetic data generation for sensors may be a challenging task. However, data normalization techniques may simplify the task, as the data generator then only has to mimic the normalized data. To further reduce the dependence on an exact data representation, our solution matches the hypotheses parameters $h$ during the hypothesis matching, and afterwards, only the binary activation matrices $B$ are compared to each other. Hence, it does not harm the solution, if the synthetic data does not perfectly match the real-world sensor data.

\section{Experimental Results}
The real-world data are multivariate time-series data covering a period of two years. 
The objects to be identified are aircraft, of which only the gears and engines are detectable by magnetometers. 
The real-world data has a high level of noise and latent factors that may reduce the data quality. Further, the data is sparse as many aircraft types occur rarely compared to a few types, which are often observed. On the other hand, the synthetic data (generated by forward modeling) has a larger aircraft type variety due to simulations of more aircraft types. The simulations were performed according to the background knowledge available. That is, for each aircraft type that passed the considered sensor field during the two-year observation period and whose aircraft dimension is known, motions were simulated with different motion vector assignments: The sensor field was passed in two directions at varying speeds 
and with small variations in the motion angles, such that aircraft do not always exactly follow the guiding line.    
All experiments were performed on a computer with OS Ubuntu 22 with 32GB memory, an Intel Core i7-4770K CPU and a GeForce GTX 1080 TI GPU. The GPU was used for the ML algorithms in the model comparison section, which are also a part of the inverse modeling of CavePerception.  

\begin{figure}[t!]
    \centering
        \includegraphics[width=0.49\textwidth]{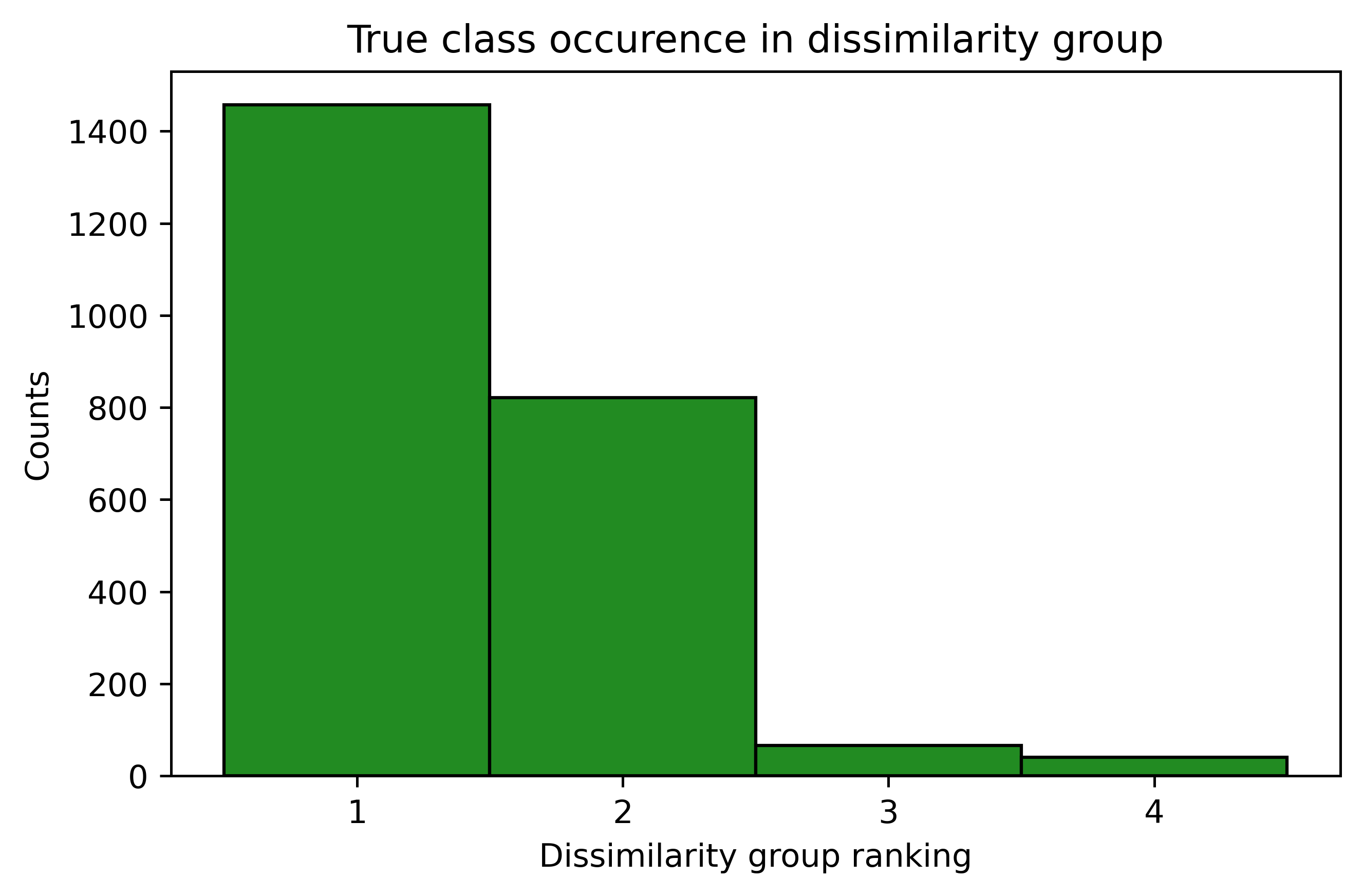}
        \includegraphics[width=0.49\textwidth]{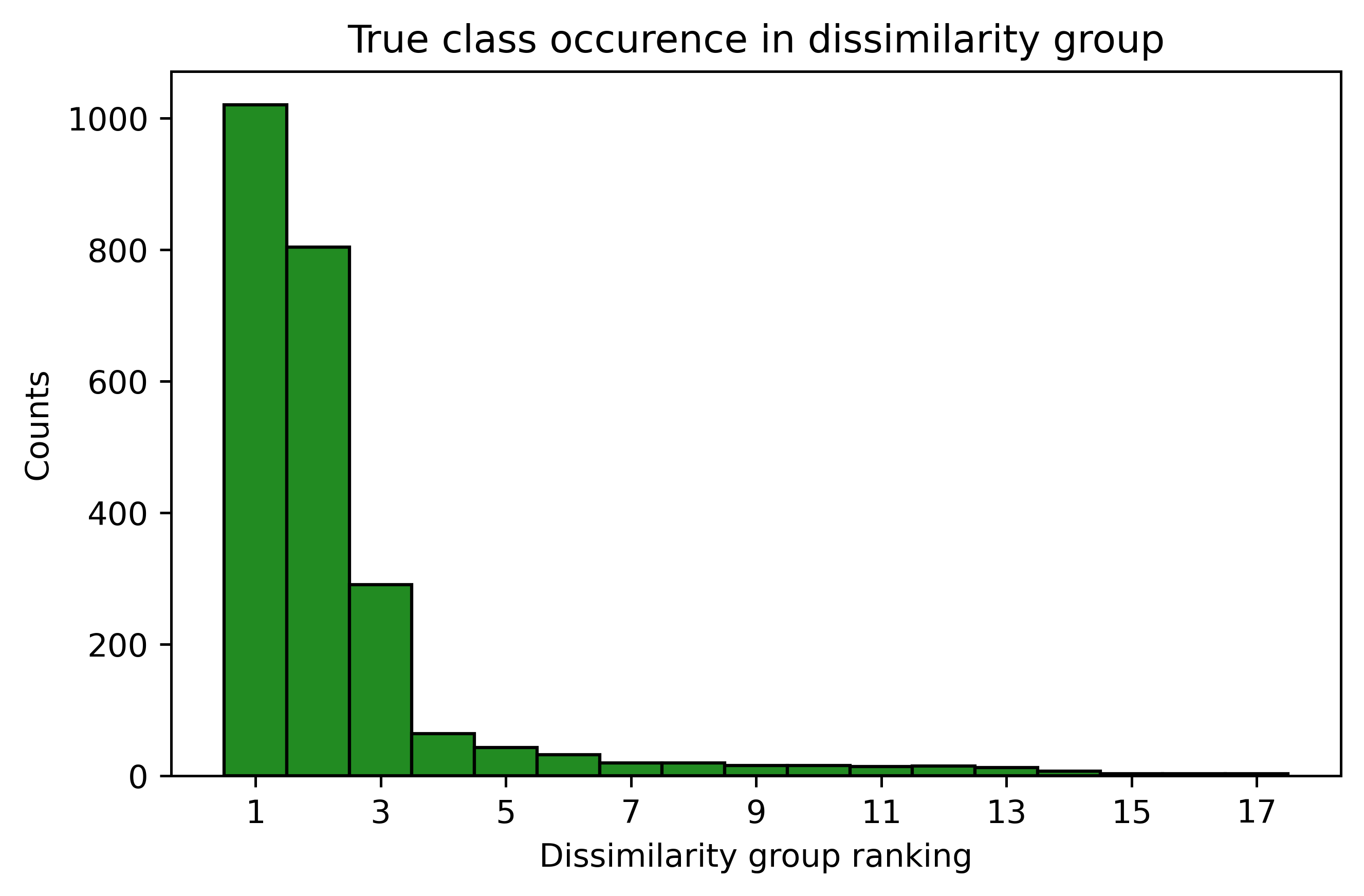}
        \includegraphics[width=0.49\textwidth]{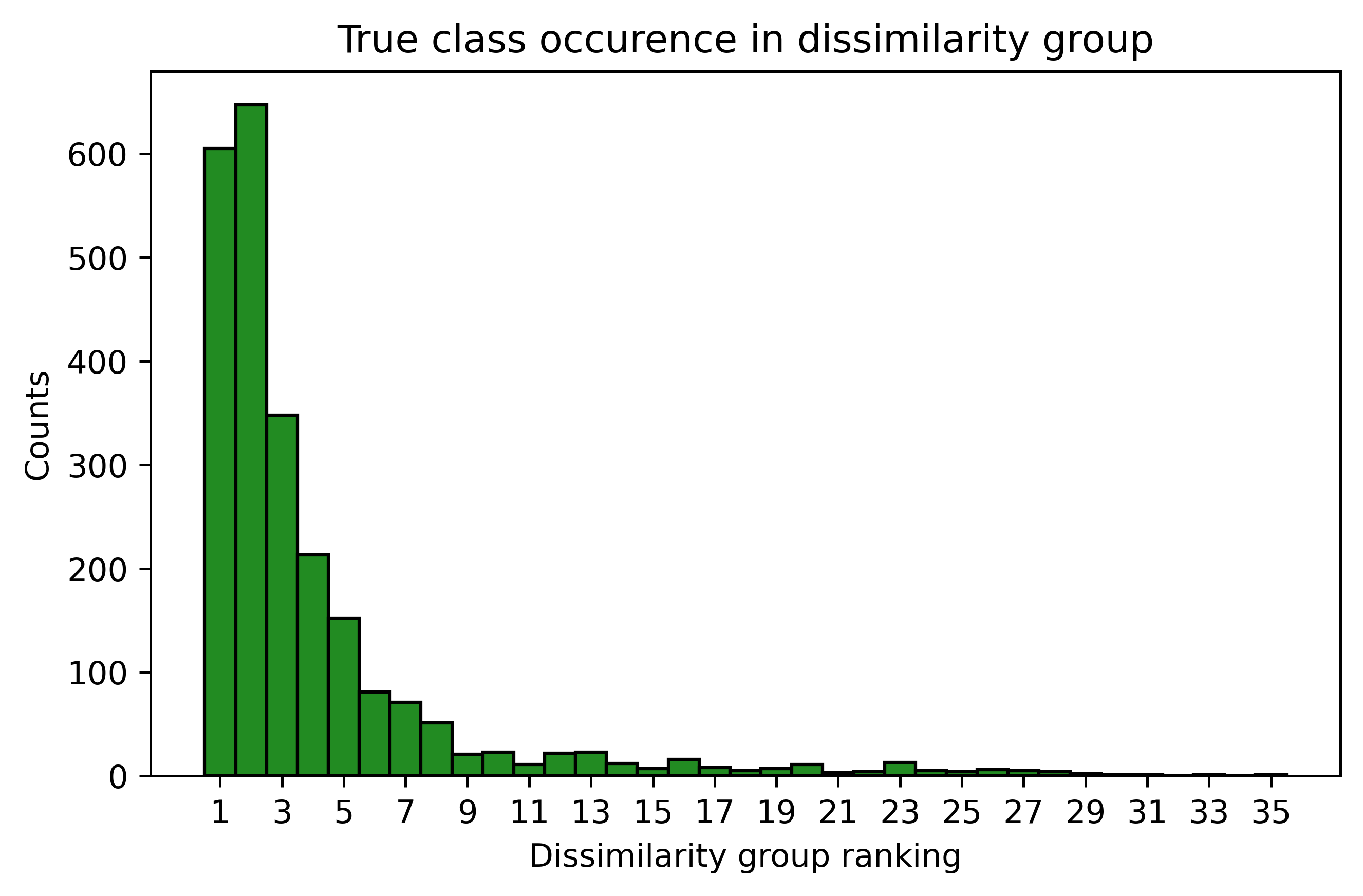}
    \caption{Group rank assignment of the true aircraft class. Group rank 1 corresponds to the predicted group of aircraft types or categories with the smallest dissimilarity. Top-Left: ARC. Top-Right: Cat. 17. Bottom: Aircraft type.}
    \label{fig:group_ranks}
\end{figure}

For the performance evaluation of CavePerception we consider three types of classification, i.e. the aircraft type, the \textit{Aerodrome Reference Code}\footnote{\href{www.skybrary.aero/articles/icao-aerodrome-reference-code}{www.skybrary.aero/articles/icao-aerodrome-reference-code}} (ARC), and cat. 17. The ARC is a universal categorization for aircraft types, introduced by the \emph{International Civil Aviation Organization} (ICAO)\footnote{\href{www.icao.int/Pages/default.aspx}{www.icao.int/Pages/default.aspx}}.
The ARC contains six categories, from A (smallest wing span) to F (largest wing span).
The aircraft type classification considers each model of aircraft separately. For example, an Airbus A320 is considered as a different type as the Airbus A320 neo, although the geometric differences are minor. Consequently, the type classification poses the most challenging classification task.
Therefore, we additionally cluster the aircraft types into 17 categories (cat. 17), where aircraft types with minor geometrical differences are grouped into one class.

\begin{table}[t]
    \caption{Real-world data results: Weighted macro averaged precision, recall, and F1-score for aircraft type, ARC, and cat. 17 classifications. RF and XGBoost are trained with 75\% of training data.}
    \label{tab:mod_comp_real}
    \centering
    \begin{tabular}{|p{1.4cm}|p{2.8cm}|p{1.6cm}|p{1.6cm}|p{1.6cm}|}
         \hline
         & & Prec. & Rec. & F1   \\
         \hline
         \multirow{5}{*}{ARC} & Random & 0.49 & 0.70 & 0.58 \\
         & RF & 0.52 & 0.66 & 0.59 \\
         & XGBoost & 0.80 & 0.67 & 0.63  \\
         & Transformer & 0.84 & 0.66 & 0.71 \\
         & CavePerception & \textbf{0.89} & \textbf{0.73} & \textbf{0.79} \\
         \hline
         \multirow{5}{1.4cm}{Cat. 17} & Random & 0.35 & \textbf{0.59} & 0.44 \\
         & RF & 0.45 & 0.54 & 0.46 \\
         & XGBoost & 0.73 & 0.55 & 0.52 \\
         & Transformer & 0.61 & 0.56 & 0.55 \\
         & CavePerception & \textbf{0.86} & 0.54 & \textbf{0.64} \\
         \hline
         \multirow{5}{1.4cm}{Type} & Random & 0.05 & 0.21 & 0.08 \\
         & RF & 0.15 & 0.00 & 0.00  \\
         & XGBoost & 0.70 & 0.08 & 0.12  \\
         & Transformer & 0.40 & 0.32 & 0.33 \\
         & CavePerception & \textbf{0.98} & \textbf{0.43} & \textbf{0.59} \\
         \hline         
    \end{tabular}
\end{table}

\paragraph{Group Rank Evaluation} 
The predicted classes of CavePerception are contained in $\tilde{\mathcal{H}}$. Based on the dissimilarities, one can obtain a ranking of dissimilarity groups $G_i=\{S_j|sdist(M_{E,S_j})=i\}$, where each group $G_i$ contains those simulations $S_j$  where the dissimilarities are equal to $i$. This approach allows us to measure the degree of deviation from the true result based on the dissimilarity values. 
Figure \ref{fig:group_ranks} shows the histograms of the dissimilarity group ranks that contain the true class. That means, group one is the group with the highest match with event $E$, while the largest group rank contains the unlikeliest matches. In these figures, we count how often the true class has been assigned to the respective group ranks. Optimally, it should always be the first group. As can be seen, predicting the ARC is the easiest task, as there are only four possible classes: $C$, $D$, $E$, and $Unknown$. The ARC values $A$, $B$, and $F$ have not been observed at this position during the time of observation, while $Unknown$ includes all those events, where the true labels of an event are unknown, i.e. vehicles, false detections, and others. Furthermore, in all three classification approaches, the true class remains in the group ranks with the smallest dissimilarities. More than 65\% of the true classes remain in the first three group ranks.
Consequently, even if less than 50\% of the correct targets were assigned to group 1 for the type and cat. 17 classifications, considering the difficulty of the classification task, the majority of the true classes are close to the highest matches with the observations. 

\paragraph{Model Comparison}
We also compare CavePerception to commonly used ML algorithms to validate the benefit of combining forward with inverse modeling procedures. Therefore, we consider Transformers, which are frequently considered one of the best performing learning algorithms, XGBoost, random forest, and a random guessing baseline, which predicts the majority class. 
The disadvantage of many standard implementations of tree-based learners is the requirement of having all data preloaded, such that the models can be trained on the whole training data at once. Due to the size of our training data and the memory size required to train these models, random forests and XGBoost were trained on 75\% of the training data. Incremental learning or similar techniques may be used to train the models on the whole dataset by iteratively updating from subsamples, however, this may introduce other challenges like representation bias from successive subsamples or the loss of correlation information. 
Hyperparameter optimization has been performed for all models to ensure a fair comparison. 
Table \ref{tab:mod_comp_real} shows that CavePerception is able to outperform all other models for each classification task, except of the recall for cat. 17, where random guessing is leading due to overpredicting the majority class. 
Notice that despite the difficulty of the aircraft type classification, CavePerception manages to achieve an impressive F1-score compared to the inverse models.  

\section{Conclusion}
In real-world applications, temporal data from sensor networks are frequently sparse and noisy. In this paper, we have shown that the combination of forward and inverse modeling can improve the overall results in such settings.
Experiments have shown that our framework, CavePerception, does not only outperform other methods on complex data, but also that its group ranking approach based on hypothesis matching is able to capture most of the correct classes within the first two to three ranked groups. The model comparison also shows that classification results may be improved by including forwardly modeled hypotheses, compared to models that are solely inversely trained. Another advantage of our solution is that missing information from inverse modeling 
is filled with assignments from synthetic simulations based on the event matching with the real-world observation. 

CavePerception should work, as far as possible, without homogeneity assumptions in real-world sensor network deployments. It is able to handle any two-dimensional sensor layout structure and also works independent of the sensor sampling frequency, as the temporal differences between consecutive points are considered instead of fixed frequency-based time points. However, it requires at least two sensor lines perpendicular to each other, of which one is necessary to estimate the motion vector, while the other one is important for classification.  
Another limitation of CavePerception is the necessity of object type geometries. If the geometry of an object type is unknown, a simulation for this type cannot be performed, resulting in a non-detection of the correct type, in case it occurs in an observation. Further, the performance of CavePerception depends on the inverse modeling results. If the inverse process yields a false classification, then it will affect the final outcome of the hypotheses matching. Therefore, it is crucial to optimize the inverse modeling procedure. In future work, this dependence on inverse modeling may be loosened by providing probabilistic results, such that the matching between inverse and forward modeling may take the probabilities into account.  
On our roadmap, CavePerception shall be further developed and integrated into a prototype for real-time traffic surveillance on airport aprons.

\begin{credits}
\subsubsection{\ackname} The authors thank the Federal Ministry for Economic Affairs and Climate Action and the German Aerospace Center for supporting the project.

\subsubsection{\discintname}
The authors have no competing interests to declare that are
relevant to the content of this article.
\end{credits}
%
%
%
%

\bibliographystyle{splncs04}
\bibliography{references}
\end{document}